%% file: ids_dm_cre_elsevier.tex
\documentclass[preprint,authoryear,12pt]{elsarticle}

\usepackage{amssymb}

\usepackage[nodots]{numcompress}

\usepackage{url}

\usepackage{graphicx}
\graphicspath{{./}{pics/}{pics/old/}{pics/6494_201306}}

\usepackage{multirow}
\usepackage{rotating}

\usepackage[cmex10]{amsmath}
\usepackage{amsfonts}
\usepackage{algorithmic}
\usepackage[algosection,algoruled,vlined]{algorithm2e}

\journal{arXiv.org}

\begin{document}


\begin{frontmatter}

\title{Anomaly Detection Framework Using Rule Extraction for Efficient Intrusion Detection}

\author{Antti Juvonen\corref{corr}}
\cortext[corr]{Corresponding author. Department of Mathematical Information Technology, P.O.~Box 35 (Agora), FI-40014 University of Jyv\"{a}skyl\"{a}, Finland. Tel.\ +358 40 357 3875.}
\ead{antti.k.a.juvonen@jyu.fi}
\author{Tuomo Sipola}
\ead{tuomo.sipola@jyu.fi}

\address{Department of Mathematical Information Technology, University of Jyv\"{a}skyl\"{a}, Finland}

\begin{abstract}
\input{abstract.tex}
\end{abstract}


\end{frontmatter}


\input{introduction.tex}

\input{related.tex}

\input{method.tex}

\input{results_iris.tex}

\input{results_kdd.tex}

\input{results_log.tex}

\input{conclusion.tex}

\section*{Acknowledgment}

\noindent This research was partially supported by the Foundation of Nokia Corporation and the Finnish Foundation for Technology Promotion. Thanks are extended to Kilosoft Oy and Pardco Group Oy. The authors thank Professors Amir Averbuch, Timo H\"{a}m\"{a}l\"{a}inen and Tapani Ristaniemi for support and guidance. 

\bibliographystyle{model4-names}
\bibliography{ids_dm_cre}


\end{document}

%% file: abstract.tex
Huge datasets in cyber security, such as network traffic logs, can be analyzed using machine learning and data mining methods. However, the amount of collected data is increasing, which makes analysis more difficult. 
Many machine learning methods have not been designed for big datasets, and consequently are slow and difficult to understand. 
We address the issue of efficient network traffic classification by creating an intrusion detection framework that applies dimensionality reduction and conjunctive rule extraction. The system can perform unsupervised anomaly detection and use this information to create conjunctive rules that classify huge amounts of traffic in real time.
We test the implemented system with the widely used KDD Cup 99 dataset and real-world network logs to confirm that the performance is satisfactory. This system is transparent and does not work like a black box, making it intuitive for domain experts, such as network administrators.

%% file: introduction.tex
\section{Introduction}







\noindent Cyber security has become a very important topic in the past years as computer networks, services and systems face new threats from attackers, which has lead to increased interest towards these matters from companies and governments. Intrusion detection systems (IDS) are an important part of cyber security. They detect intrusions and abnormal behavior in networks or other systems producing big data. Practical environments are always different, and this affects the choice of the IDS and it's detection algorithms \citep{MOLINA2012,LIAO2013}. These systems usually apply one of two detection principles: signature-based or anomaly-based~\citep{SCARFONE2007}. Signature-based approach means using manually created rules that detect intrusions, whereas anomaly-based systems try to profile normal behavior and detect abnormal action dynamically. It is also possible to combine these approaches to form a hybrid IDS, as we have done in this paper. This means creating signatures automatically and they can be periodically updated. In addition to the previous methodologies, stateful protocol analysis can be applied \citep{LIAO2013}. This means finding unexpected sequences of commands. Moreover, anomaly detection systems could be used in combination with existing next-generation firewalls and other systems, so that the IDS benefits from the best properties of both approaches. Many different algorithms can be used for anomaly detection, e.g. self-organizing maps \citep{RAMADAS2003} and support vector machines \citep{TRAN2004}. Another approach is to use genetic algorithms in intrusion detection context \citep{TAN2012,GOYAL2012}.

\begin{figure}[t]
\centering
\includegraphics[width=8.5cm]{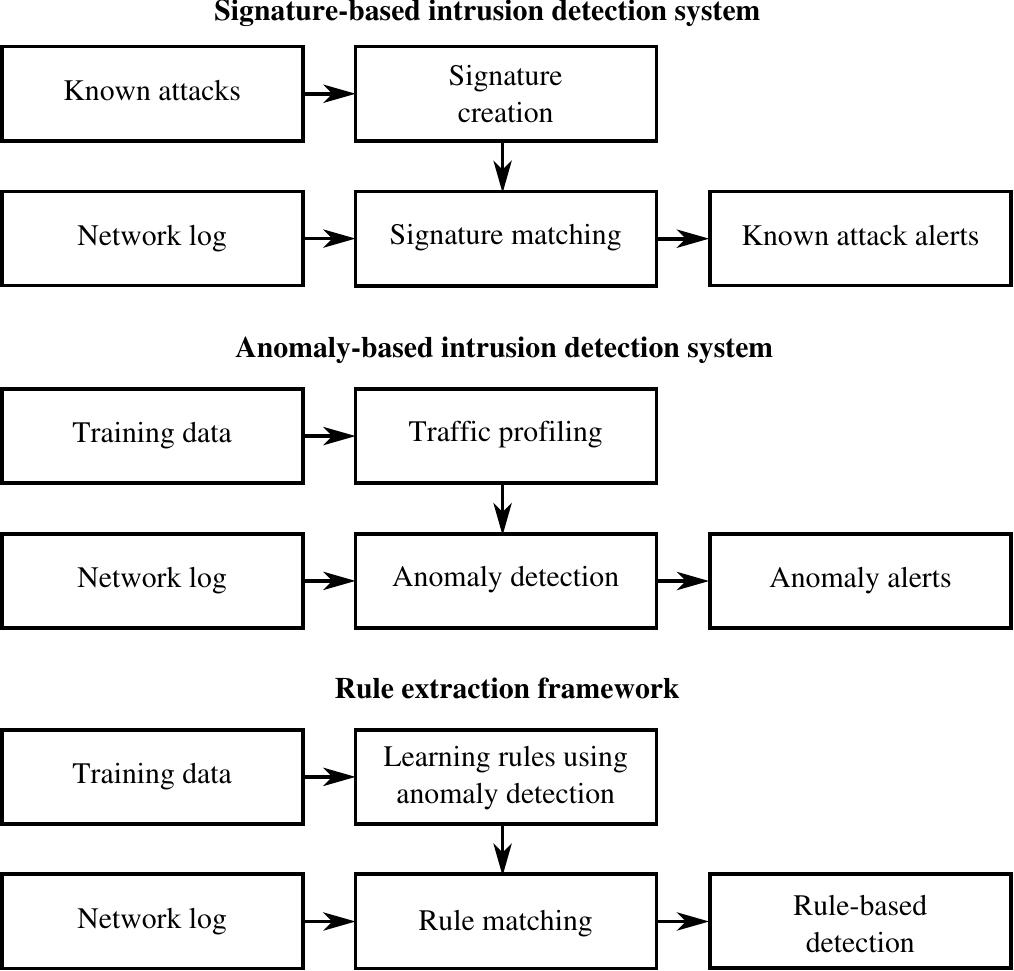}
\caption{Different IDS principles.}
\label{fig:idss}
\end{figure}

Machine learning offers many benefits for intrusion detection \citep{CHANDOLA2009}. However, it has limits that should be considered. 
Semantic gain, i.e.\ the practical meaning of the results, is usually more valuable in practical cases than marginal increases in performance accuracy \citep{SOMMER2010}. Anomaly detection should be used in combination with existing systems to bring added value. One problem with machine learning algorithms used for anomaly detection is the fact that many of them work like a black box from the end user perspective. It is not an easy task to know how the algorithm internally functions, and because of this, companies have difficulties deploying these systems. To overcome this problem, rule extraction algorithms have been proposed \citep{CRAVEN1994}. These algorithms aim to create rules that replicate or approximate classification results. The main benefit of deploying rule-based systems is that they are fast and they might reveal comprehensible information to the users better than black box algorithms. However, it seems that many rule extraction algorithms depend on neural networks~\citep{HUYSMANS2006}. Furthermore, rule extraction itself is usually a supervised learning task and needs previously created classification and labeling information in order to work.

We propose an intrusion detection framework that uses diffusion map methodology for dimensionality reduction and clustering to automatically label network traffic data. Figure \ref{fig:idss} shows the different IDS principles, and how the rule extraction framework relates to them. Our system follows the three overall stages of network IDS: data parametrization, training stage and detection stage \citep{GARCIA2009}. The learning phase to label the traffic is unsupervised. Subsequently, the classification information is forwarded to a rule extraction algorithm that creates conjunctive rules. These automatically generated rules are used to classify new incoming traffic data. The rules can also be analyzed by a domain expert in order to acquire new information about the nature of the data. The presented framework combines knowledge-based expert system and machine learning-based clustering approaches \citep{GARCIA2009}. Consequently, the black box nature of some of the existing algorithms is avoided. The framework does not need preceding information about the intrusions that are present in the training data. In addition, the structure of the system is dynamic and individual algorithms can be changed if necessary. Low throughput and high cost, as well as lack of appropriate metrics and assessment methodologies have been identified as common problems in anomaly-based IDSes \citep{GARCIA2009}. 

This paper contributes to methodology and practical analysis in the field of network security. Firstly, the framework addresses the above issues by generating rules that can classify incoming traffic with low computational cost. Secondly, this research uses a well-known public dataset and appropriate metrics to assess the performance. Finally, real-world network data is used to further investigate the effectiveness of the framework in a practical situation. The main contribution of this paper is finalizing and detailed description of the framework \citep{JUVONEN2013} and tests with more varied data that show better detection rates than before. 

The paper is structured as follows. First, we go through related reseach concerning dimensionality reduction methodologies and rule extraction. In the methodology section the overall system, all of the used algorithms and performance metrics are described. Then, we introduce our experimental cases and present the obtained results. Finally, we discuss the benefits and limitations of the system as well as future research directions.

%% file: related.tex
\section{Related work}

\noindent This section briefly discusses research that is related to the methods used in this article. There are five areas that are covered: dimensionality reduction, anomaly detection, rule extraction, big data approaches and the earlier frameworks designed by the authors. These points have contributed to the design of the current framework. 

Firstly, dimensionality reduction has been widely researched in the intrusion detection context. Perhaps the most well-known method is principal component analysis (PCA) \citep{JOLLIFFE2005}, which has been used in network anomaly detection \citep{RINGBERG2007,CALLEGARI2011}. However, it has some problems, such as the fact that it cannot handle non-linear data. 
In general, manifold learning approaches try to learn the structure of the data, retaining some meaningful qualities of the data mining problem. The point is to contain most of the information in the data using fewer dimensions. The goals in such a setting are understanding and classification of data and generalization for use with new data \citep{LEE2007}. Several manifold learning methods have been used for intrusion detection, including Isomap and locally linear embedding (LLE) \citep{ZHENG2009a,ZHENG2009b,YUANCHENG2010}.
This paper uses the diffusion map manifold learning method, which is a non-linear dimensionality reduction method \citep{COIFMAN2006a}. Diffusion map meth\-odology has been used for network traffic classification and SQL intrusion detection \citep{DAVID2009,DAVID2010,DAVID2011}. We have previously used it for anomaly detection from network logs \citep{SIPOLA2011,SIPOLA2012}. This framework is enhanced by using clustering to detect multiple behaviors in the data \citep{JUVONEN2012}. In these experiments, the diffusion map algorithm acts like a black box, which is a drawback that makes the system hard to understand for people who are not familiar with data mining technologies.

Secondly, related research on anomaly detection in intrusion detection context is explored. Some of the common general approaches to intrusion detection are statistical methods, machine learning and data mining \citep{PATCHA2007}. Statistical methods do not require prior knowledge about attacks, but they need a certain statistical distribution, which is not always the case. Machine learning based methods (e.g., Bayesian networks and principal component analysis (PCA)) aim to learn from the behavior and improve the performance over time. Advantages and drawbacks depend on the used algorithm, e.g., PCA cannot handle non-linear data, which is why we use diffusion map methodology for dimensionality reduction. Finally, data mining methods such as genetic algorithms and artificial neural networks attempt to find patterns and deviations automatically from the data. It is also possible to use some kind of hybrid system. In addition, anomaly detection techniques can be divided into classification-based, nearest neighbor-based and clustering-based methods \citep{CHANDOLA2009}. Also, artificial immune systems have been used extensively in intrusion detection \citep{KIM2007}. Ensemble systems have also been successful in the area \cite{MUKKAMALA2005}. Using computational and artificial intelligence methods makes it possible to create adaptive, fault tolerant and fast systems \citep{WU2010,LIAO2013}. Our system combines many of the previously mentioned approaches such as clustering-based anomaly detection and dimensionality reduction. The system, its differences to existing methodologies and our contributions are explained in more detail at the end of this section.

Thirdly, rule extraction is considered. In order to understand the black-box nature of non-linear classifiers, rule extraction methods can create sets of rules that describe the behavior of such systems in a more understandable manner \citep{MARTENS2008}. Our implementation of the presented framework is based on the conjunctive rule extraction algorithm \citep{CRAVEN1994}. However, the overall framework does not depend on any specific clustering or classification algorithm. The choice of the used classification method influences the selection of other algorithms in the framework, because of varying performance and robustness. The approach presented in this research resembles spectral clustering frameworks and can be though as a special case of them \citep{BACH2006,LUXBURG2007}.
There are many algorithms that extract rules from trained neural networks, such as TREPAN \citep{CRAVEN1996} and Re-RX \citep{SETIONO2008}. These algorithms depend on the neural network architecture and the associated weights.
Another common approach is to use support vector machines as a basis for rule extraction \citep{NUNEZ2002,BARAKAT2004,BARAKAT2010}. One example for generating signatures for detecting polymorphic worms is Polygraph, which uses disjoint content substrings to identify them \citep{NEWSOME2005}. Regular expressions generated with a supervised domain expert dataset have also been used to screen unwanted traffic \citep{PRASSE2012}. 
In addition, many swarm intelligence algorithms can be used for different intrusion detection applications. For example, ant colony optimization (ACO) can be utilized to detect intrusions, detect the origin of an attack and also induction of classification rules \citep{KOLIAS2011}. The field of swarm intelligence contains many other algorithms that are useful for IDS purposes, such as particle swarm optimization (PSO) and ant colony clustering (ACC).

Next, many of the above mentioned algorithms can be scaled up for big data applications using various parallel and distributed approaches \citep{BEKKERMAN2012}. For example, the map-reduce framework has proved itself as a feasible method for machine learning \citep{CHU2006}. The $k$-means algorithm used in our system can also be parallelized using map-reduce \citep{ZHAO2009}. Moreover, the traditional intrusion detection methods can be sped up using graphics processors \citep{VASILIADIS2008}. 

Finally, the authors have already proposed a system for extracting conjunctive rules from HTTP network log from real-life web servers \citep{JUVONEN2013}. The data mining approach to network security is similar to the methodology of this paper. We now extend and improve this framework for different kinds of data. Our framework works in an unsupervised manner and does not depend on the selection of the used classification algorithm.

The system proposed and tested in this paper combines many of the approaches mentioned previously in this section. The dimensionality reduction phase could have been done using principal component analysis, but it sometimes fails with non-linear data. We have learned through experiments that diffusion map sometimes gives more accurate results. In addition, clustering based anomaly detection has also been used previously in the literature. However, our approach is different because we do not cluster the original data, but the data points in low-dimensional space after diffusion map phase. The diffusion map already helps to separate the points, reducing the error that $k$-means clustering might introduce. Finally, because the diffusion map is not computationally very efficient, we use conjunctive rule extraction to approximate the clustering results and classify data. This way we aim to combine accurate results and feasible computational speed.


%% file: method.tex
\section{Methodology}

\begin{figure}[t]
\centering
\includegraphics[width=6cm]{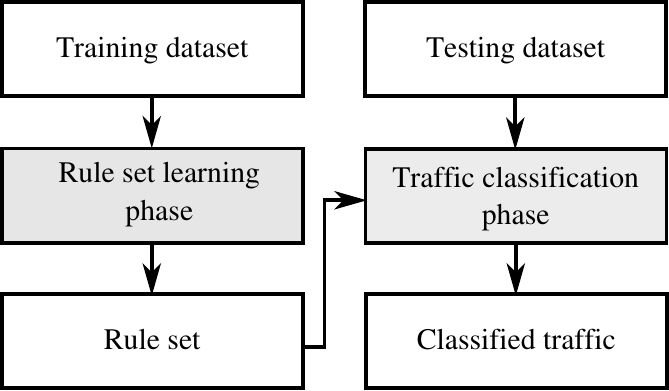}
\caption{Block diagram of the overall approach.}
\label{fig:process}
\end{figure}

\noindent Our system is divided into two phases: training phase and testing phase. Training consists of preprocessing and ruleset learning, and as an end result it produces a ruleset that can be used for traffic classification. Testing phase uses the created rules to classify testing data. Performance of the testing phase is also measured. 
The overall approach is described in Figure \ref{fig:process}.

The training phase consists of the following steps:

\begin{itemize}
\item Training data selection
\item Feature extraction
  \begin{itemize}
  \item Discretization of continuous data (binning)
  \item Binarization
  \end{itemize}
\item Unsupervised learning
  \begin{itemize}
  \item Dimensionality reduction
  \item Clustering
  \end{itemize}
\item Detecting normal traffic
\item Rule extraction
\end{itemize}

\begin{figure}[t]
\centering
\includegraphics[width=7cm]{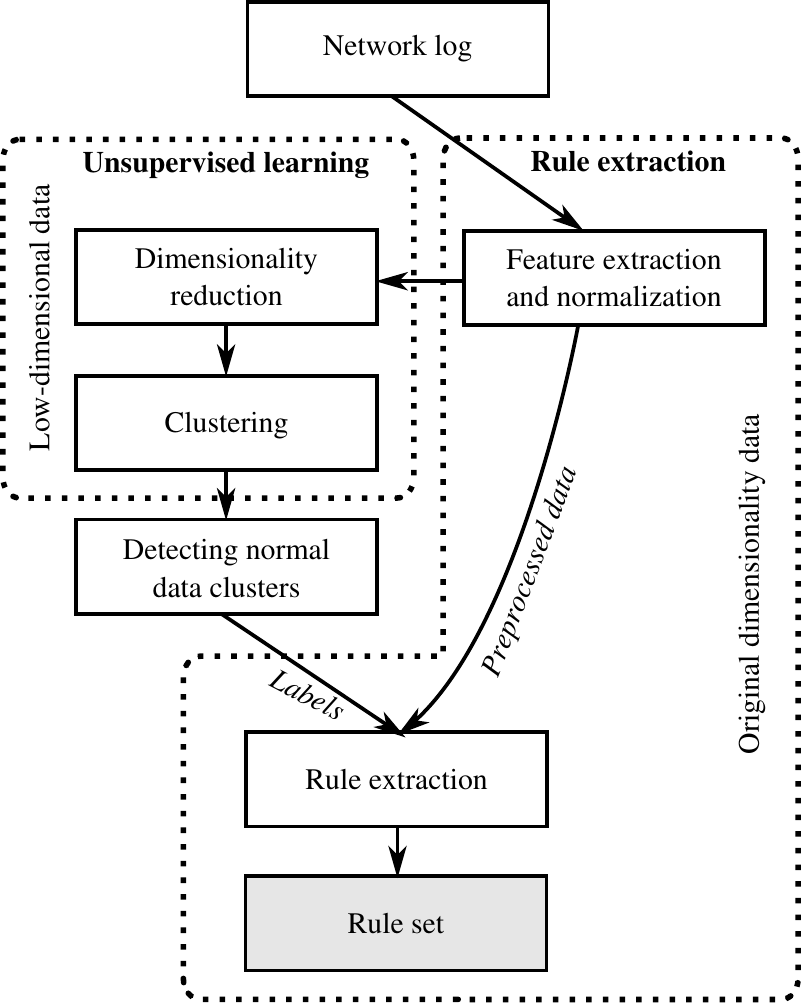}
\caption{Block diagram of the ruleset learning process.}
\label{fig:architecture}
\end{figure}

Figure \ref{fig:architecture} shows this ruleset learning phase in detail. Previously unknown data (network log) is used as an input, and this data is automatically labeled in an unsupervised manner. However, detecting the normal data cluster might need expert input. This classification information can then be used to create the ruleset, which tries to simplify the process of dimensionality reduction and clustering to human-readable rules.

Testing, or ruleset matching phase can be summarized as follows:

\begin{itemize}
\item Test data feature extraction
\item Rule matching
\end{itemize}

This phase preprocesses new data and uses the rules created in the learning phase to classify the testing data as normal or intrusive. If needed, classification into more than two classes is also possible if this is taken into account in the training phase. The new traffic that the system identifies as normal is naturally not flagged. Traffic corresponding to intrusion rules are flagged as attacks. This flagging resembles misuse detection. In addition, traffic that does not match any rules, is flagged as unknown anomalies. 

\subsection{Dimensionality reduction using diffusion map} \label{subsec:dimred}

\noindent Dimensionality of the feature space is reduced using diffusion maps. This training produces a low-dimensional model of the data, which facilitates clustering and identification of internal structure of the data. Diffusion maps are useful in finding non-linear dependencies. At the same time the diffusion distances in the initial feature space correspond proportionally to the Euclidean distances in the low-dimensional space \citep{COIFMAN2005,COIFMAN2006a,NADLER2006}. 

Each data point is represented by a feature vector $x_i \in \mathbb{R}^n$. The whole dataset $X = \{ x_1,\ x_2,\ x_3,\ \dots\ x_N \}$ is a collection of these feature vectors. To perform the diffusion map, an affinity matrix 

\begin{equation*}
W(x_i, x_j) = \exp \left( {\frac{-||x_i - x_j||^2}{\epsilon}} \right)
\end{equation*}

\noindent describing the similarity between the data points is calculated. Here affinity is defined using the Gaussian kernel. The parameter $\epsilon$, which is responsible for the neighborhood size, is selected from the optimal region in the weight matrix sum 

\begin{equation*}
L = \sum_{i=1}^{N} \sum_{j=1}^{N} W_{i,j},
\end{equation*}

\noindent which can be plotted on a logarithmic scale for the identification of the middle region \citep{COIFMAN2008}, while $\epsilon$ is changed. 

The row sums of $W$ are collected to the diagonal of matrix $D$. This matrix is used to normalize $W$ in order to create transition matrix

\begin{equation*}
P = D^{-1}W.
\end{equation*}

This transition matrix is symmetrized as 

\begin{equation*}
\tilde{P} = D^{\frac{1}{2}}PD^{-\frac{1}{2}} = D^{-\frac{1}{2}}WD^{-\frac{1}{2}}. 
\end{equation*}

The singular value decomposition (SVD) of $\tilde{P}$ finds the eigenvectors $v_k$ and eigenvalues $\lambda_k$, which can be used to construct low-dimensional coordinates for the data points. Each data point in the original data gets corresponding diffusion map coordinates:

\begin{equation*}
x_i \to [ \lambda_1 v_1(x_i),\ \lambda_2 v_2(x_i) \dots\ \lambda_d v_d(x_i) ].
\end{equation*}

The first eigenvectors retain most of the information in the data, which is why the later eigenvectors are left out. Some information is lost but the error is bounded and lower dimensionality facilitates clustering. 

\subsection{Clustering} \label{subsec:cluster}

\noindent The $k$-means method groups the data points into clusters. The algorithm is widely used in data mining and its implementation is simple. It finds the centroid point of clusters and then assigns each point to the nearest cluster centroid. This process is iterated until the clustering does not change after updating centroids. The algorithm and use cases are described in more detail in literature \citep{JAIN1988,TAN2006,JAIN2010}. We chose to use $k$-means clustering together with diffusion map because its use is justified in the literature \citep{LAFON2006}.

In order to use $k$-means, the number of clusters needs to be determined. The clustering is repeated many times and the quality of the obtained clusters is measured. For $i$th data point, silhouette 

\begin{equation*}
s(i) = \frac{b(i)-a(i)}{max\{a(i),b(i)\}}
\end{equation*}

\noindent expresses the quality of the clustering. Here $a(i)$ is the average dissimilarity of the point to all the other points in the same cluster, and $b(i)$ is the smallest of the average dissimilarities to all the other clusters.  The larger the silhouette value, the better the clustering is for the data point. The average of the silhouette values for all data points is used to evaluate the quality of clustering. These qualities are used to compare different numbers of clusters \citep{ROUSSEEUW1987}. 

\subsection{Rule extraction} \label{subsec:re}

\noindent A rule defines a number of easy-to-understand conditions for the features in order to classify the data points according to those conditions. A ruleset consists of many rules and replicates a certain classification result. In an ideal situation a rule should be understandable by network experts without the effort of knowing the underlying classifiers. Rules correspond to data features, and are useful as such for finding the root cause of intrusions or other anomalous behavior. 

It is not feasible to find all the possible rules because they span such a huge space. In an optimal setting all the rules should be checked, but in practice a sub-optimal but efficient method is used. Such rule extraction systems have been used to learn the information provided by, e.g., neural networks \citep{CRAVEN1994,RYMAN2010} or support vector machines \citep{NUNEZ2002,BARAKAT2004,BARAKAT2010}. 

There are two main approaches to rule extraction. Firstly, \emph{decompositional} rule extraction algorithms exploit the internal mathematical or algorithmic features of the underlying data mining method \citep{DAVILA2001}. Secondly, \emph{pedagogical} algorithms do not depend on the method used \citep{CRAVEN1994}. They only take into account the input data and classification result. We take the pedagogical approach, because this makes it possible to use different kinds of algorithms for data classification.

A conjunctive rule is a logical expression that combines logical symbols using the conjunction (i.e.\ AND) operation. Such a symbol states whether the value of a binary feature must be true or false. The absence of the symbol means that the feature can be anything. Note that in this research a binary feature means a truth value either belonging to a symbolic category, or a truth value that tells in which bin a continuous value belongs to. 

Let us consider an example where we have five features $a$, $b$, $c$, $d$, $e$. This means that the feature matrix contains five columns corresponding to each binary feature. Each data point corresponds to a row in the matrix. Here is an example ruleset containing three rules classifying data into two separate classes $c_1$ and $c_2$:

\begin{equation*}
\begin{aligned}
r_1 = & \neg a & \text{ for class } c_1 \text{,} \\
r_2 = & a \land b \land c \land \neg d \land e & \text{ for class } c_1 \text{,} \\
r_3 = & a \land b \land \neg c & \text{ for class } c_2 \text{.} 
\end{aligned}
\end{equation*}

In this case, rule $r_1$ means that if the value of the feature $a$ is false for a single data point, it belongs to class $c_1$. The values for the rest of features do not matter. Other rules work in a similar way: rule $r_3$ states that $a$ and  $b$ have to be true while $c$ must be false. If this rule $r_3$ holds true, the point is classified to class $c_2$. 

In our actual implementation the truth values are converted to 1 and $-1$, and the values that do not matter are expressed as 0. Therefore, the rule $r_1$ can be expressed using vector 
$\begin{pmatrix} -1 & 0 & 0 & 0 & 0 \end{pmatrix}$. This approach makes rule matching easy from the computational perspective.

The conjunctive rule algorithm (\ref{alg:conjunctive}) creates a ruleset for the training data. It creates new rules until the whole dataset is covered without misclassification \citep{CRAVEN1994}. If the value of a symbol in a rule does not affect the classification result using the rules, the symbol can be omitted. The final rules contain only the symbols that are essential for the classification. The rules generated using the algorithm can be used for simple matching in the testing phase. 

\begin{algorithm}
\caption{Conjunctive rule extraction \citep{CRAVEN1994}.}
\label{alg:conjunctive}
\begin{algorithmic}
\REQUIRE data points $E$, classes $C$
\ENSURE rules $R_c$ that cover $E$ with classification $C$
\REPEAT 
  \STATE $e$ := get new training observation from $E$
  \STATE $c$ := get the classification of $e$ from $C$
  \IF{$e$ not covered by the rules  $R_c$}
    \STATE $r$ := use $e$ as basis for new rule $r$
    \FORALL{symbols $s_i$ in $r$}
      \STATE $r'$ = $r$ with symbol $s_i$ dropped
      \IF{all instances covered by $r$ are of the same\\ class as $e$}
        \STATE $r$ := $r'$
      \ENDIF
    \ENDFOR
    \STATE add rule $r$ to the ruleset $R_c$
  \ENDIF
\UNTIL{all training data analyzed}
\end{algorithmic}
\end{algorithm}

\subsection{Performance metrics}

\noindent After classifying data points as normal or intrusive, and running the testing phase, the following performance metrics are calculated.
Table \ref{tab:model_cm} shows the confusion matrix and some abbreviations used in this research. Note that correct alarms of intrusions are regarded as true positive identifications.

\input{model_cm.tex}

Sensitivity or true positive rate (TPR) tells how well intrusive data points are identified:
\begin{equation*}
\mathit{sensitivity} = \frac{tp}{tp + fn}.
\end{equation*}

False positive rate (FPR) is calculated in a similar way. False positives are data points that are classified as attacks but are actually normal. In intrusion detection context this is often called false alarm rate:
\begin{equation*}
\mathit{fpr} = \frac{fp}{fp + tn}.
\end{equation*}

Specificity or true negative rate (TNR) explains how well normal data points are found:
\begin{equation*}
\mathit{specificity} = \frac{tn}{fp + tn} = 1 - fpr.
\end{equation*}

Accuracy tells how many correct results there are compared to the whole data:
\begin{equation*}
\mathit{accuracy} = \frac{tp + tn}{tp + fp + fn + tn}.
\end{equation*}

Precision explains the proportion of true positive classifications in all positive results, i.e.\ how many alarms were correct and not false alarms:
\begin{equation*}
\mathit{precision} = \frac{tp}{tp + fp}.
\end{equation*}

The previous metrics are extensively used but cannot alone describe the results in a satisfactory way. For this reason, we also use Matthews correlation coefficient \citep{BALDI2000,BROWN2006,FAWCETT2006}. It can measure the quality of binary classification and is regarded as a good metric for measuring the overall classification performance:
\begin{equation*}
\mathit{MCC} = \frac{tp \times tn - fp \times fn}{\sqrt{(tp+fp)(tp+fn)(tn+fp)(tn+fn)}}.
\end{equation*}

%% file: model_cm.tex

\begin{table}[!t]

\small
\renewcommand{\arraystretch}{2.0}
\caption{Confusion matrix.}
\label{tab:model_cm}

\vspace{0.5em}
\centering
\begin{tabular}{c c | c c}

& & \multicolumn{2}{c}{\textbf{predicted}} \\
& &  normal & attack \\
\hline
\multirow{2}{*}{\rotatebox{90}{\hspace{-5pt}\textbf{actual}}} & normal  &     true negative (tn) &      false positive (fp) \\
& attack &     false negative (fn) &    true positive (tp)

\end{tabular}

\end{table}

%% file: results_iris.tex
\section{Iris data: an illustrative example} \label{sec:iris}

\noindent In this section a toy problem is presented to clarify certain concepts in the framework. The rule extraction algorithm described in subsection \ref{subsec:re} is illustrated here using a practical example. In machine learning community, the IRIS dataset \citep{ASUNCION2010} is commonly used to illustrate the algorithms in a simple case. We use this well-known dataset to demonstrate the rule generation process and what kind of information is produced as a result. The data contains 150 samples from 3 different species of flowers. There are four different measurements made from each flower: sepal length and width as well as petal length and width. All these features are continuous. The task is to cluster the data and find corresponding rules using our framework.

Feature generation is performed by first dividing all 4 measurements into 3 bins. Larger number of bins is possible but in this case it was not necessary. The binning process works in the following way. For this dataset, the minimum value of petal width measurement is 0.1 cm and maximum value 2.5 cm. For this feature, the boundaries of the bins are as follows:

\begin{center}
\setlength\unitlength{2pt}
\begin{picture}(100,17)
\put(0,7){\line(1,0){100}}
\multiput(12,7)(25,0){4}{\line(0,1){5}}
\put(9,2){$0.1$}
\put(34,2){$0.9$}
\put(59,2){$1.7$}
\put(84,2){$2.5$}
\put(18,8){$Bin1$}
\put(44,8){$Bin2$}
\put(68,8){$Bin3$}
\end{picture}
\end{center}

Other features are binned using the same principle. After binning, the values are binarized from multi-category features to single-category binary features for rule extraction. Since 3 bins are used for each of the 4 features, we get 12 binary columns in the processed matrix. Consequently, the final feature matrix is of size $150 \times 12$. These 12 binary values also correspond to the rules, so that each rule has 12 elements.

The data is clustered into 3 groups using the $k$-means clustering algorithm. There is no dimensionality reduction step beforehand because 12-dimensional feature space is still rather low-dimensional. The clustering is used to label flower samples using 3 labels. Subsequently, the rules are extracted. 

\begin{figure}[t]
\centering
\includegraphics[width=8cm]{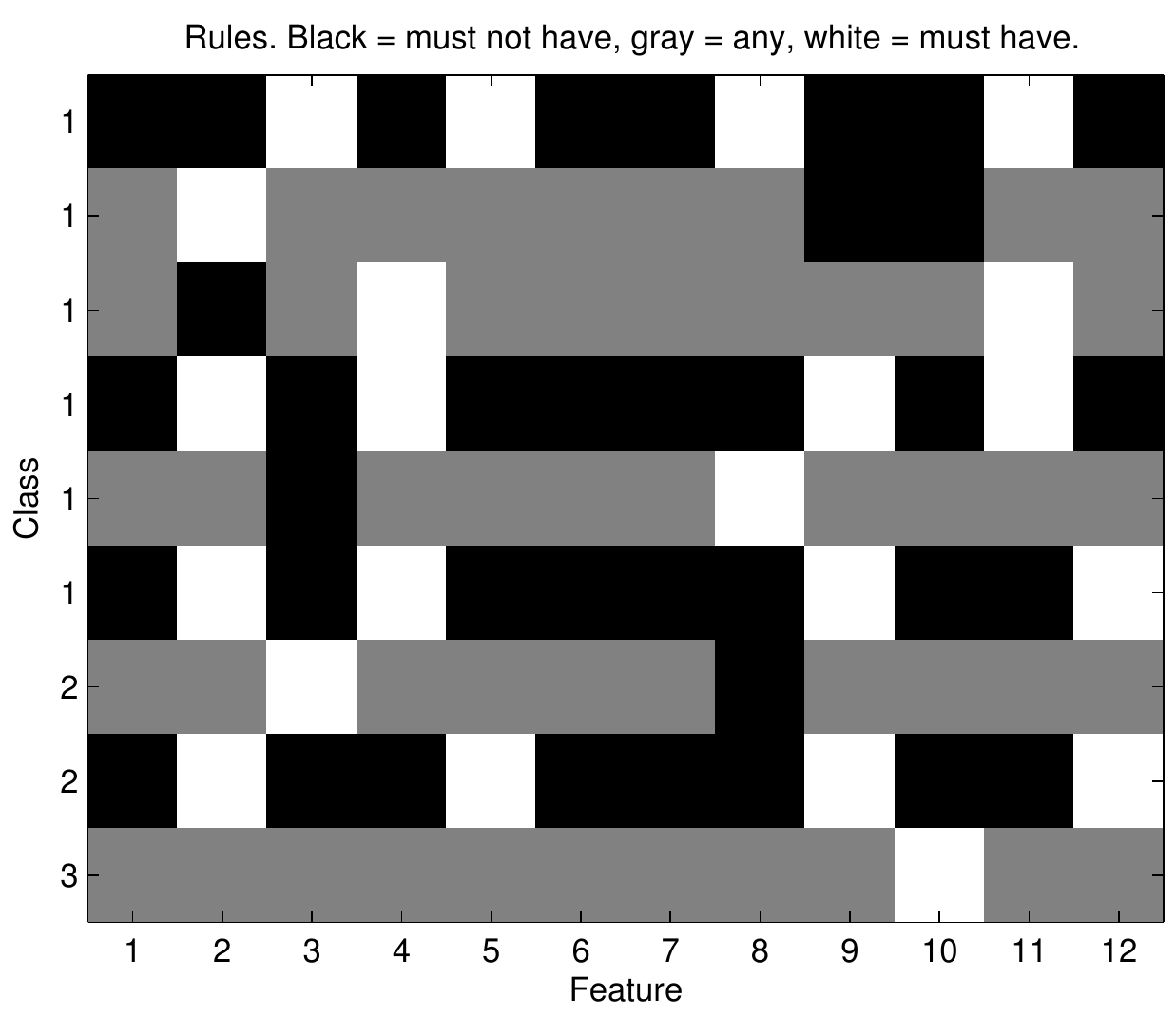}
\caption{Rules generated from clustered IRIS data. Each line represents one rule. Note that classes 1, 2 and 3 correspond to the three species.}
\label{fig:iris_rules}
\end{figure}

The resulting ruleset can be seen in Figure \ref{fig:iris_rules}. There are 9 rules in total. There is only one rule generated for class 3 (\textit{setosa} species). The rule states that if the fourth measurement (petal width) belongs to bin 1, the flower's species is \textit{setosa}. Bin 1 means that the petal width must be between 0.1 and 0.9 cm, as demonstrated above. The rules with mostly 'must not have' (i.e.\ black) symbols are usually created for single data points, which might be outliers. This is an example of how information can be gained from the rules. We can see that the classification process no longer works like a black box, but is based on human-readable information.

%% file: results_kdd.tex
\section{KDD Cup 99 data} \label{sec:kdd}

\noindent For this experiment we used KDD Cup 99 10\% data \citep{ASUNCION2010}. It has some known limitations \citep{MAHONEY2003,SABHNANI2004,TAVALLAEE2009} stemming from its source, the likewise problematic DARPA dataset \citep{MCHUGH2000}, but it is publicly available and labeled, which makes it easy for researchers to use the data and compare results \citep{DAVIS2011}. A serious concern is the unrealistic distribution of intrusions. Therefore, it is common to use the smaller 10\% dataset or a custom version of it  \citep{KOLIAS2011}. The dataset was created by gathering data from a LAN that was targeted with multiple attacks. It contains 494,021 traffic samples and each sample includes 41 features. The features include both continuous and symbolic values. The attacks fall into four general categories: denial-of-service, remote-to-local, user-to-root and probing. There are 24 attack categories in total in the training data. Data features include things such as duration of the connection, protocol type, network service, transferred bytes and failed login attempts. Even though the actual labels (normal or intrusive) are known, this information is not used in the rule creation phase. We use unsupervised learning for the framework, and only use the actual labels for getting the performance metrics and evaluating the results.

\subsection{Feature extraction}


\noindent The data that we use for testing and evaluation contains both continuous and discrete values. This must be taken into account in the feature extraction phase. In network traffic context, features such as the number of bytes transferred during a connection are continuous, and, e.g., used protocol type in contrast is a discrete valued feature.


During the preprocessing phase, continuous feature values must be transformed into discrete ones for the used rule extraction algorithm. Symbolic discrete values can be left alone at this stage. Discretization is performed using binning. We do this by acquiring the maximum and minimum values of each continuous variable, and dividing the measurements into $n$ bins using equal intervals. The number of bins can be chosen according to the situation, e.g. $n=3$ or $n=10$. There is a practical example in section \ref{sec:iris}. Let us now present another example, where we assume that we have 3 bins for each variable and there are 3 different variables ($f1, f2, f3$) and 3 data points (rows). The number of the chosen bin for each measurement is placed in a feature matrix:

\begin{center}
\begin{tabular}{c | c | c}
\textbf{f1} & \textbf{f2} & \textbf{f3} \\
\hline
1 & 2 & 1 \\
3 & 2 & 3 \\
2 & 1 & 1
\end{tabular}
\end{center}

This kind of matrix is not yet usable for rule extraction. In the final preprocessing phase, we binarize the matrix. This gives us a matrix that uses the following format:

\begin{center}
\begin{tabular}{c c c | c c | c c}
\textbf{f1=1} & \textbf{f1=2} & \textbf{f1=3} & \textbf{f2=1} & \textbf{f2=2} & \textbf{f3=1} & \textbf{f3=3}\\
\hline
1 & 0 & 0 & 0 & 1  & 1  & 0\\
0 & 0 & 1 & 0 & 1  & 0  & 1\\
0 & 1 & 0 & 1 & 0  & 1  & 0
\end{tabular}
\end{center}

For example, the first column f1=1 means that if the chosen bin for the first variable is 1 on a certain row, the binary value in this binarized matrix will be 1, otherwise 0. In other words, a column tells whether a specific variable measurement belongs to a certain bin or not. This way we get nominal values for all the features. Columns that would be all zeros can be omitted. Now the matrix is in a format that can be directly used with the rule extraction algorithm. In addition, the rule extraction requires labeling information. The process of acquiring this is explained in the following subsections.

\subsection{Random sampling from the 10\% KDD set}

\noindent For initial testing to showcase the feasibility of the approach, 5,000 data lines are randomly selected for testing and 5,000 for training, totaling 10,000 lines. The purpose of this initial experiment is to see if a rather limited training sample can represent the whole data. The feature matrix is binned and binarized as described previously. After this, dimensionality is reduced using a diffusion map.

\begin{figure}[t]
\centering
\includegraphics[width=8cm]{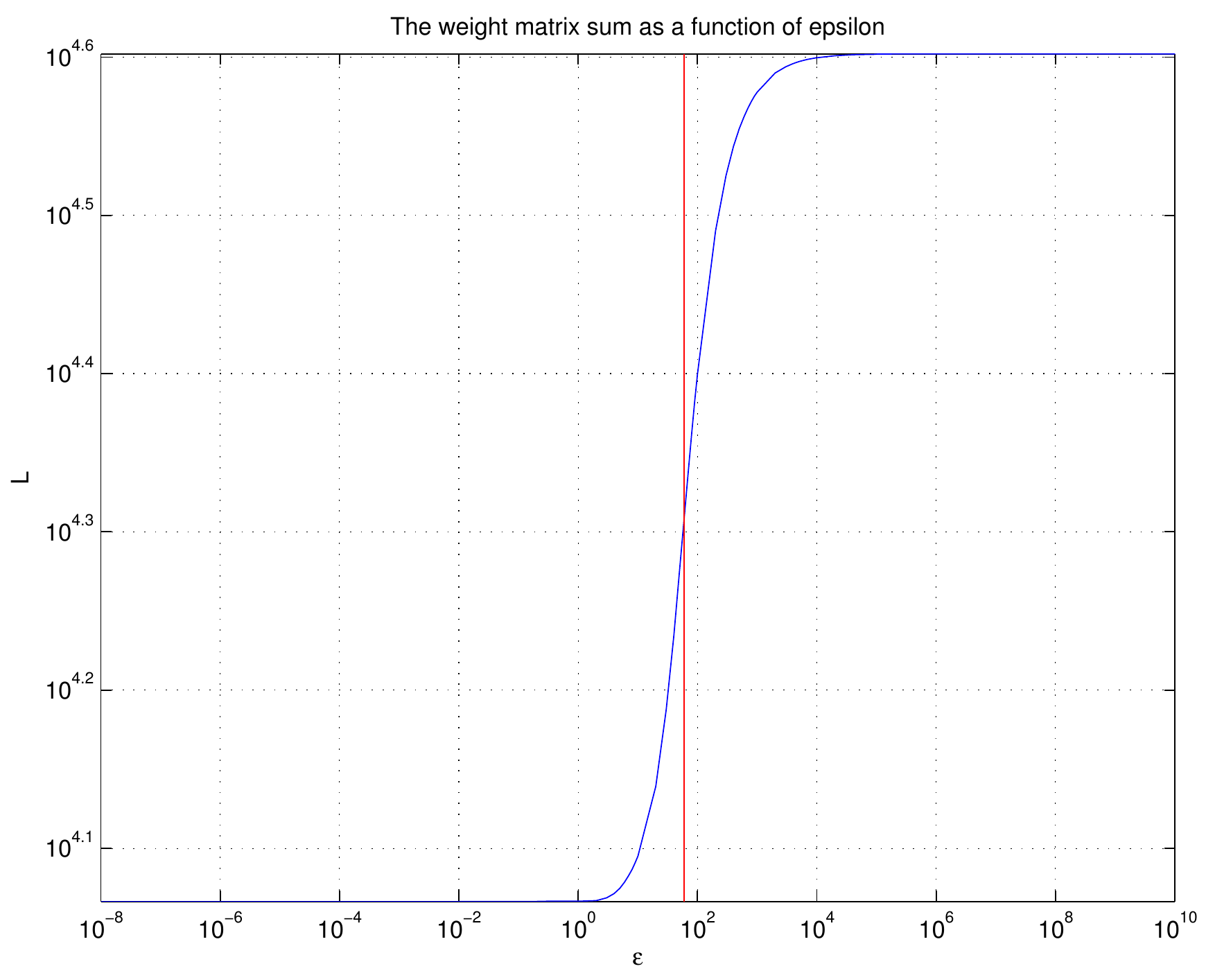}
\caption{Finding the optimal region for epsilon selection for the diffusion map algorithm.}
\label{fig:epsilon}
\end{figure}

In order to perform the dimensionality reduction in an optimal way, the value for parameter $\epsilon$ must be determined. Selecting $\epsilon$ for the diffusion map is performed by selecting the value from the region between the extremes of a log-log plot, as seen in Figure~\ref{fig:epsilon}. The quality $L$ is calculated by summing the weight matrix, as explained in subsection \ref{subsec:dimred}. For each value of $\epsilon$, the sum is calculated using 200 randomly selected samples from the training data. 

\begin{figure}[t]
\centering
\includegraphics[width=8cm]{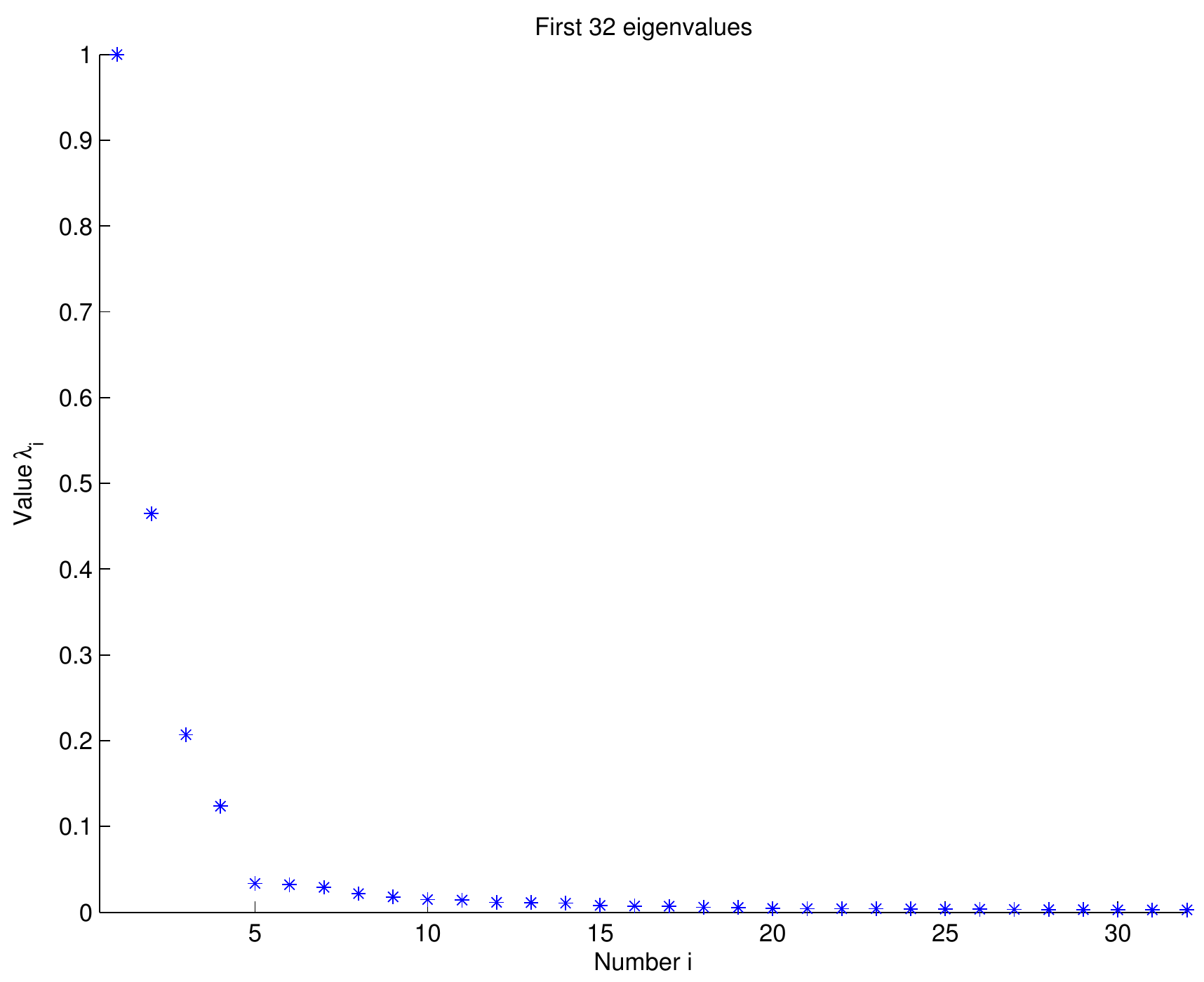}
\caption{Eigenvalues of the diffusion map algorithm. The eigengap is visible between 4th and 5th eigenvalue.}
\label{fig:eigenvalues}
\end{figure}

The eigenvalues $\lambda_k$ from the diffusion map method are presented in Figure~\ref{fig:eigenvalues}. The first eigenvalue is always $1$ and is therefore ignored. Next three are used in the analysis, since they seem to cover most of the data. The eigengap is visible after them. The rest of the eigenvalues contain little information and are discarded. The underlying assumption of the analysis is that the first few eigenvalues and eigenvectors contain the most useful clustering information.

\begin{figure}[t]
\centering
\includegraphics[width=8cm]{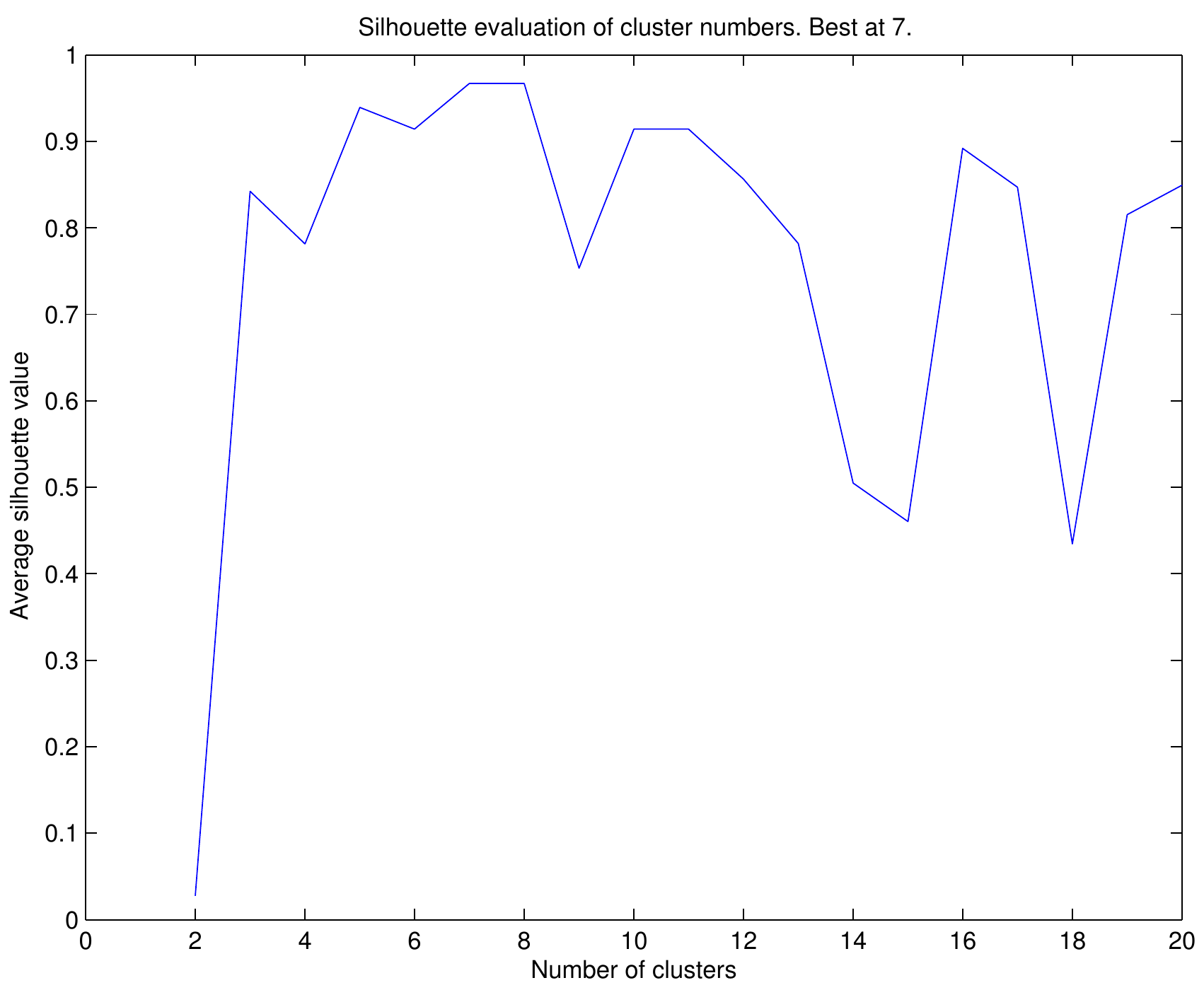}
\caption{Cluster number quality measured using silhouette.}
\label{fig:cluster_goodness}
\end{figure}

After dimensionality reduction the data is clustered using $k$-means algorithm. This step is believed to separate the normal data points from the attacks. 
To evaluate the best number of clusters, clustering process is repeated with different number of clusters, from 2 to 20. Figure \ref{fig:cluster_goodness} shows the quality of clustering with the average silhouette metric. Out of the silhouette values, the highest one should be selected. Based on this metric, the data is divided into 7 clusters.

\begin{figure}[t]
\centering
\includegraphics[width=8cm]{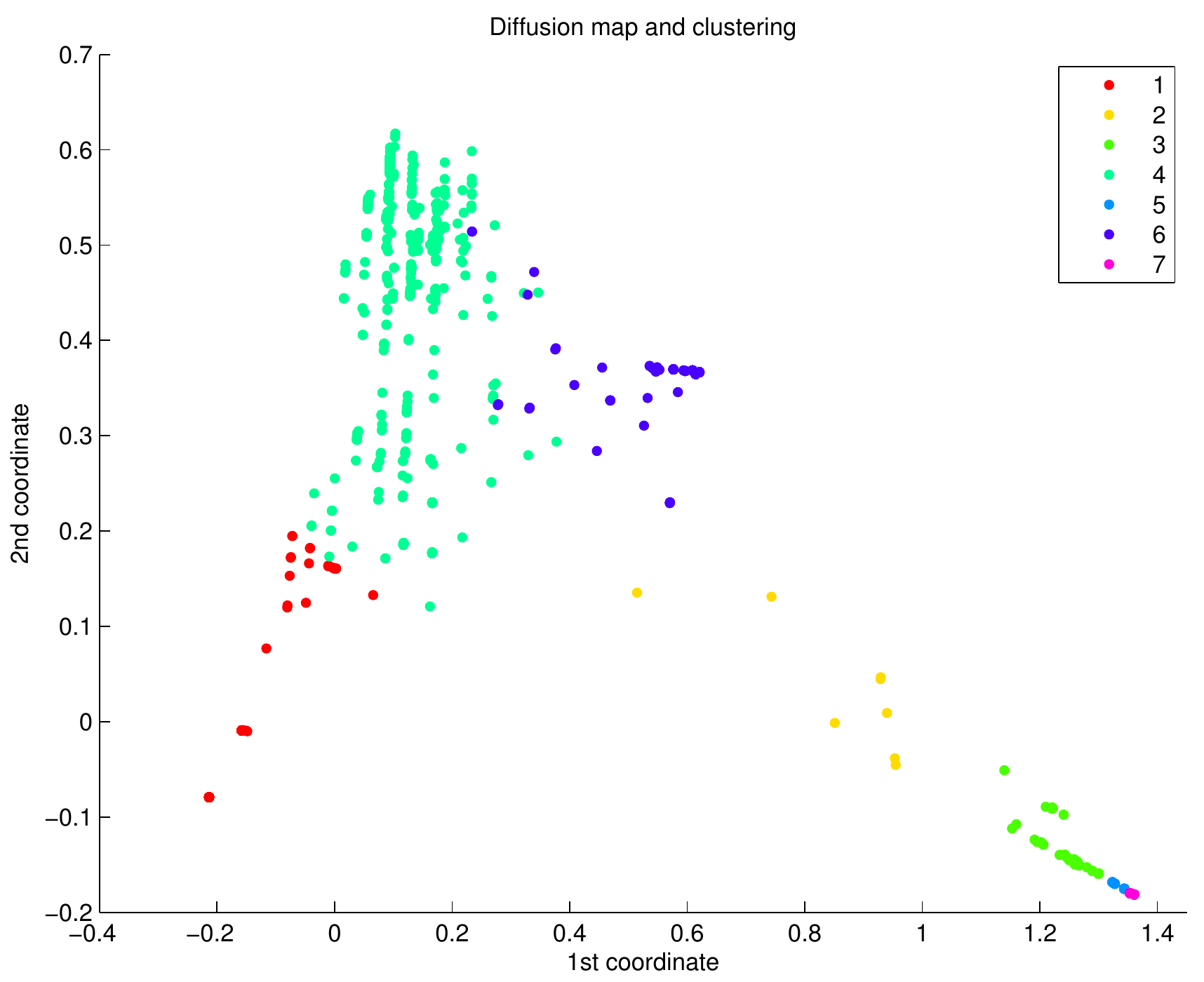}
\caption{Found clusters mapped to the two first coordinates in the low-dimensional space.}
\label{fig:dm_clusters}
\end{figure}

Figure \ref{fig:dm_clusters} shows the obtained clusters after dimensionality reduction and $k$-means clustering. The normal data is assumed to lie in cluster number 4, and all the others are regarded as attacks for labeling purposes. Normally we would assume that the largest cluster is the one containing normal data points, meaning that whole process could be completed automatically. However, due to the fact that KDD Cup 99 dataset unnaturally contains more intrusions than actual normal traffic, we had to choose the normal cluster using manual inspection.

\begin{figure}[t]
\centering
\includegraphics[width=8cm]{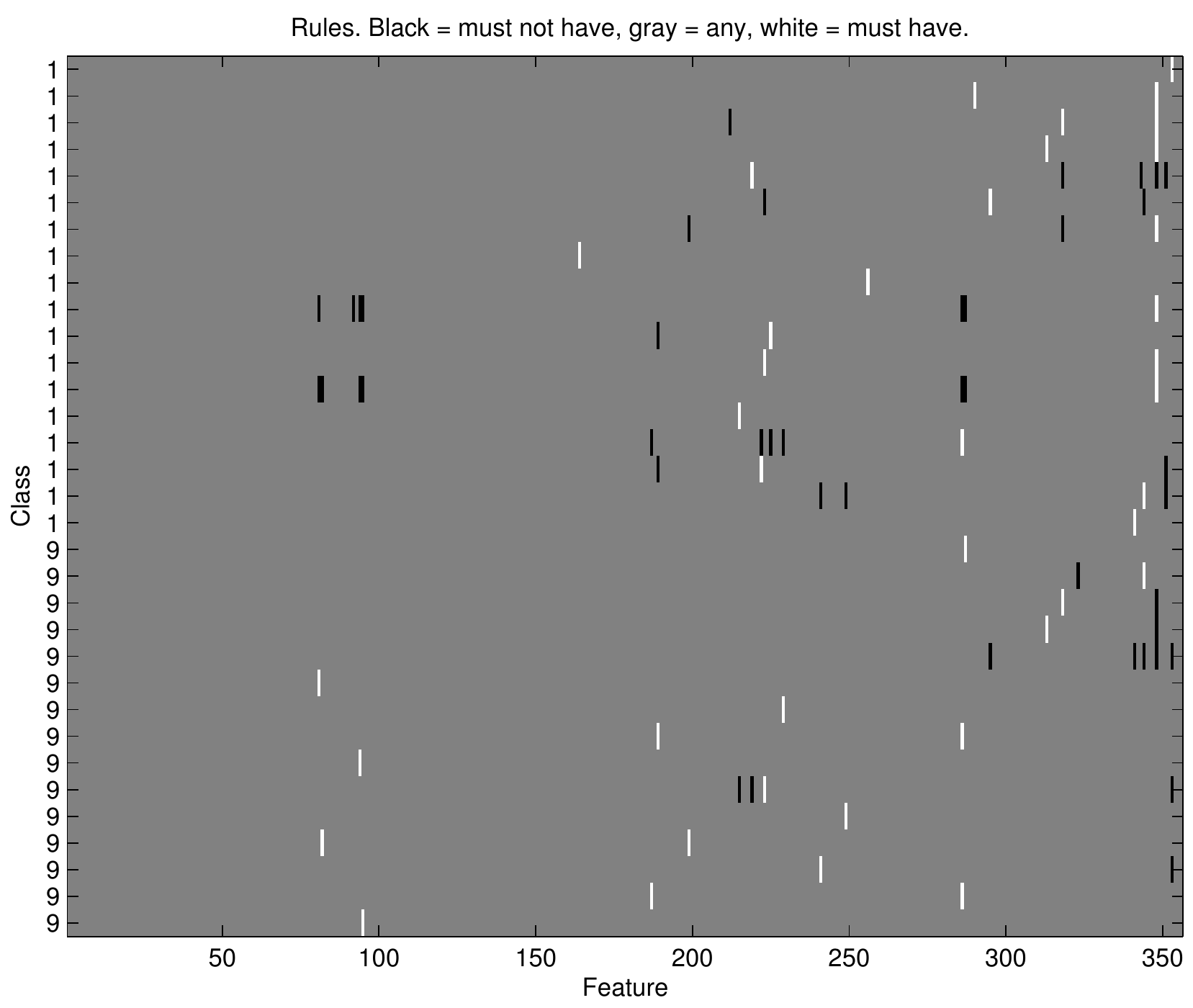}
\caption{Conjunctive rules. Each line represents one rule. Note that class 1 means normal and class 9 intrusive.}
\label{fig:rules}
\end{figure}

The clustering is assumed to represent the classification of the training dataset. This information is used to create labeling, which is needed for rule creation phase. This is an example of how the rule extraction algorithm can be used in an unsupervised manner, even though it is in principle based on supervised learning. The conjunctive rules for this dataset are illustrated in Figure \ref{fig:rules}. There are 18 rules for the normal class and 15 rules for the attack class, 33 in total. This shows that the data can be represented and classified by a relatively small and simple set of rules.

Table \ref{tab:5000_acc} shows the performance metrics for the 5,000 training lines setup. The confusion matrix is shown in Table \ref{tab:5000_cm}. From these it can be seen that we get a good intrusion detection accuracy with a relatively low number of false alarms. It is important to note that Matthews correlation coefficient usually gives lower values than other commonly used metrics. Using the created rules to classify new incoming traffic is very fast, which is the main benefit of this method. Furthermore, the rules reveal comprehensible information about normal and intrusive data points. 

\input{10_5000_5000_accuracy.tex}

\input{10_5000_5000_cm.tex}

\subsection{The whole 10\% KDD set}

\noindent For a more comprehensive experiment we use the whole KDD Cup 99 10\% dataset. A randomly selected subset of 5\% (24,701 data points) is used for training phase, leaving 469,320 data points for testing. The goal is to use small training set size compared to the testing set to alleviate the scaling of the diffusion map method. Data preprocessing, binning and binarization as well as diffusion map dimensionality reduction and clustering are done in the same way as described in the previous case. As earlier, the normal clusters had to be identified using manual inspection. Rule extraction is again performed using the conjunctive rule algorithm. 





Table \ref{tab:whole_acc} shows the performance measures for the testing data. From the table we can see that the overall performance is not as good as in the previous case. However, true positive rate is slightly better. The most important metric to look at is the Matthews correlation coefficient. The whole dataset has more variability in it and is not as compact as the smaller one. For these reasons the Matthew's correlation coefficient seems to be lower with this dataset. There are also more false alarms.
Table \ref{tab:whole_cm} shows the confusion matrix, which supports the results in the performance metrics table. Note that attacks are regarded as positive classifications. 

\input{10_whole_accuracy.tex}

\input{10_whole_cm.tex}

The obtained results confirm, using a bigger dataset, that a manifold learning framework can perform intrusion detection adequately. Earlier research supports the use of similar techniques \citep{ZHENG2009a,ZHENG2009b,YUANCHENG2010}. This also validates the assumption that a system based on conjunctive rules can achieve promising performance. Next, we will investigate real-life data. 

\subsection{Ruleset size}

\begin{figure}[t]
\centering
\includegraphics[width=8cm]{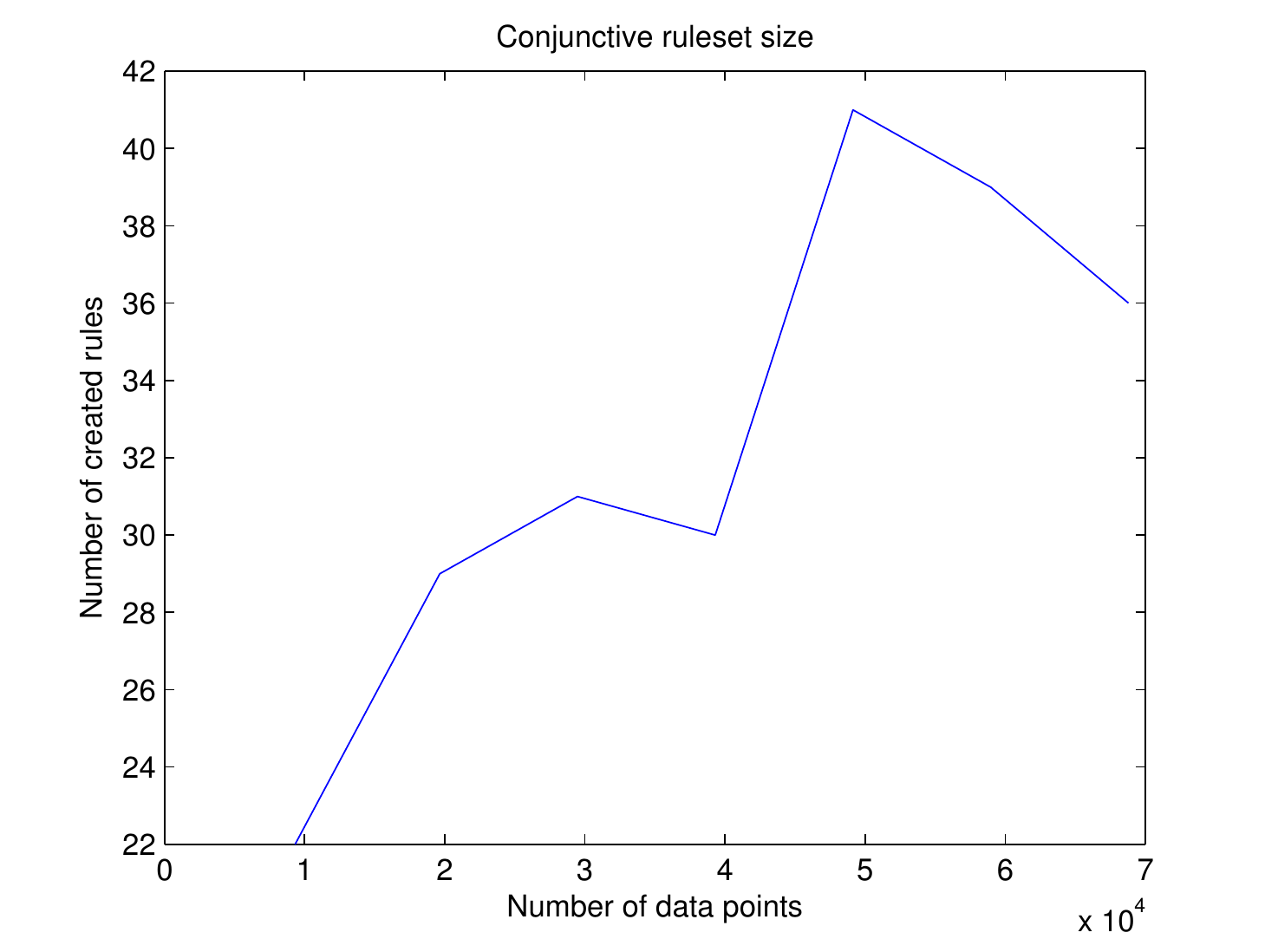}
\caption{Training data amount effect on rule set size.}
\label{fig:rssize}
\end{figure}

\noindent Training data size and characteristics affect the size of the created conjunctive ruleset. From Figure \ref{fig:rssize} we can see that the ruleset size does not increase as fast as the amount of training data points. Because the algorithm is not deterministic (random starting rule), it is also possible that the ruleset size decreases. If the data points are very similar, the ruleset might converge and additional rules are not created. This means that the ruleset size created in training phase stays manageable even with bigger training data. 

%% file: 10_5000_5000_accuracy.tex

\begin{table}[!t]

\small
\renewcommand{\arraystretch}{1.3}
\caption{Performance metrics for the 5,000 testing data lines.}
\label{tab:5000_acc}

\centering
\begin{tabular}{c c}

\textbf{Metric} & \textbf{Value \%} \\
\hline
Sensitivity (TPR)   & 98.44 \\
FPR                 & 1.25 \\
Specificity (TNR)   & 98.75 \\
Accuracy            & 98.50 \\
Precision           & 99.70 \\
Matthews corr. coef.& 95.31

\end{tabular}

\end{table}

%% file: 10_5000_5000_cm.tex

\begin{table}[!t]

\small
\renewcommand{\arraystretch}{2.0}
\caption{Confusion matrix for the 5,000 testing data lines.}
\label{tab:5000_cm}

\vspace{0.5em}
\centering
\begin{tabular}{c c | c c}

& & \multicolumn{2}{c}{\textbf{predicted}} \\
& &  normal & attack \\
\hline
\multirow{2}{*}{\rotatebox{90}{\hspace{-5pt}\textbf{actual}}} & normal  &     947 &      12 \\
& attack &      63 &    3,978 

\end{tabular}

\end{table}

%% file: 10_whole_accuracy.tex

\begin{table}[!t]

\small
\renewcommand{\arraystretch}{1.3}
\caption{Performance metrics for the whole 10\% dataset.}
\label{tab:whole_acc}

\centering
\begin{tabular}{c c}

\textbf{Metric} & \textbf{Value \%} \\
\hline
Sensitivity (TPR)   & 98.46 \\
FPR                 & 5.59 \\
Specificity (TNR)   & 94.41 \\
Accuracy            & 97.67 \\
Precision           & 98.63 \\
Matthews corr. coef.& 92.64

\end{tabular}

\end{table}

%% file: 10_whole_cm.tex

\begin{table}[!t]

\small
\renewcommand{\arraystretch}{2.0}
\caption{Confusion matrix for the whole 10\% dataset, 469,320 testing data lines.}
\label{tab:whole_cm}

\vspace{0.5em}
\centering
\begin{tabular}{c c | c c}

& & \multicolumn{2}{c}{\textbf{predicted}} \\
& &  normal & attack \\
\hline
\multirow{2}{*}{\rotatebox{90}{\hspace{-5pt}\textbf{actual}}} & normal  &     87,228 &      5,162 \\
& attack &      5,795 &    371,135 

\end{tabular}

\end{table}

%% file: results_log.tex
\section{Real-world network data} \label{sec:log}

\noindent Finally, we use real-world network server log data as a case study to describe the feasibility of the approach. This is the most realistic scenario presented in the article. This dataset does not contain labeling, and thus the diffusion map is used to cluster and classify the data for ruleset evaluation. These results have been explored earlier \citep{JUVONEN2013}. The dataset comes from a real web server, which serves the company's web services and pages. The logs are received from Apache servers running in a company network, and the lines are formatted as follows:

\begin{verbatim}
127.0.0.1 - - 
[01/January/2012:00:00:01 +0300] 
"GET /resource.php?
parameter1=value1&parameter2=value2 
HTTP/1.1" 200 2680
"http://www.address.com/webpage.html" 
"Mozilla/5.0 (SymbianOS/9.2;...)" 
\end{verbatim}

The log contains IP address, timestamp, HTTP request, server response code and browser information.

\subsection{Feature extraction}

\noindent In order to be able to use our methods to analyze the log data, textual log lines must be transformed into numerical vectors. For this purpose, we use $n$-grams \citep{DAMASHEK1995}. In this context, an $n$-gram is a substring of characters obtained by sliding window with size $n$ through the string. For example, let's assume that we have two strings, \texttt{anomaly} and \texttt{analysis}. If we use bigrams (2-grams), we get the following two feature vectors that can be placed into a matrix:

\begin{center}
   \begin{tabular}{ c  c  c  c  c  c  c  c  c  c  }
     an & no & om & ma & al & ly & na & ys & si & is \\ \hline
     1  & 1  & 1  & 1  & 1  & 1  & 0  & 0  & 0  & 0  \\ 
     1  & 0  & 0  & 0  & 1  & 1  & 1  & 1  & 1  & 1  \\
   \end{tabular}
\end{center}

For this study, bigrams are used. The frequencies of each individual bigram are stored in a feature matrix, where each row corresponds to one log line, as demonstrated in the example above. Longer $n$-grams could also be used, but due to added columns for individual $n$-grams the feature matrix becomes very large.

\subsection{Training size effect}

\noindent The first real-world dataset contains 4,292 log lines. We extract 490 unique bigrams from the data. Accordingly, after preprocessing the resulting matrix  contains 4,292 rows and 490 columns. Because there is no existing labeling, the clustering analysis produces a label for each line, which identifies it as normal of anomalous. We select randomly 2,000 data points for ruleset creation. The labeling information of these lines is used for the rule extraction process. This leaves us with 2,292 log lines that are not present during ruleset learning phase, and these lines are used as a testing set which is classified using the ruleset. Labels for testing set are only used for performance evaluation, not the classification process.

The ruleset for the first real-world dataset contains 6 rules, 2 for classifying detecting normal traffic and 4 for anomalies. Then we classify the testing set using these rules. We discover that one anomalous sample is not covered by any rule, and therefore not classified. The rest are correctly classified. All the normal traffic falls under a single rule. In this case the system works with almost 100\% accuracy, which suggests that the amount of data is limited.

\begin{figure}[t]
\centering
\includegraphics[width=8cm]{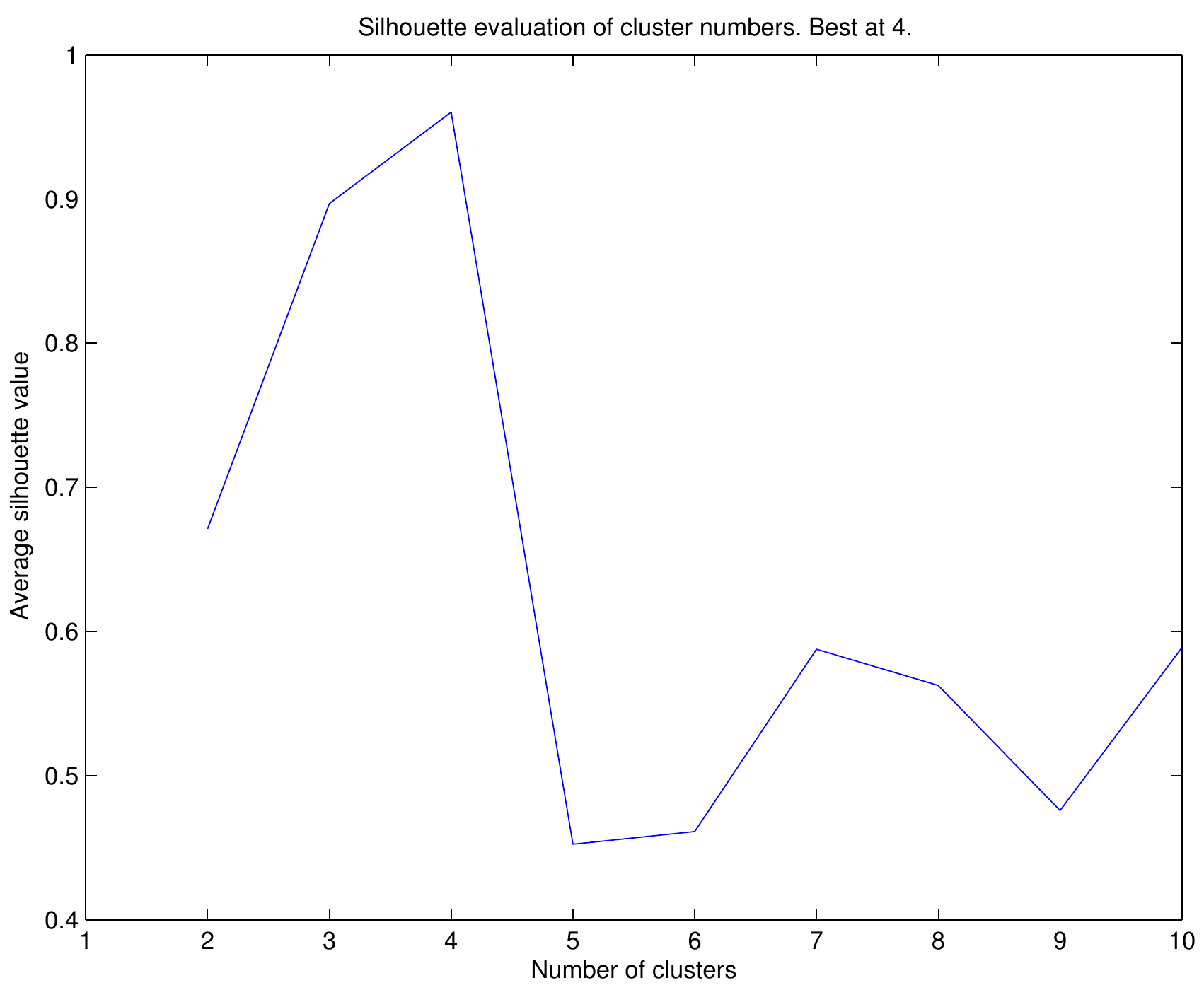}
\caption{Optimal cluster number for $k$-means.}
\label{fig:cluster_goodness_real}
\end{figure}

\begin{figure}[t]
\centering
\includegraphics[width=8cm]{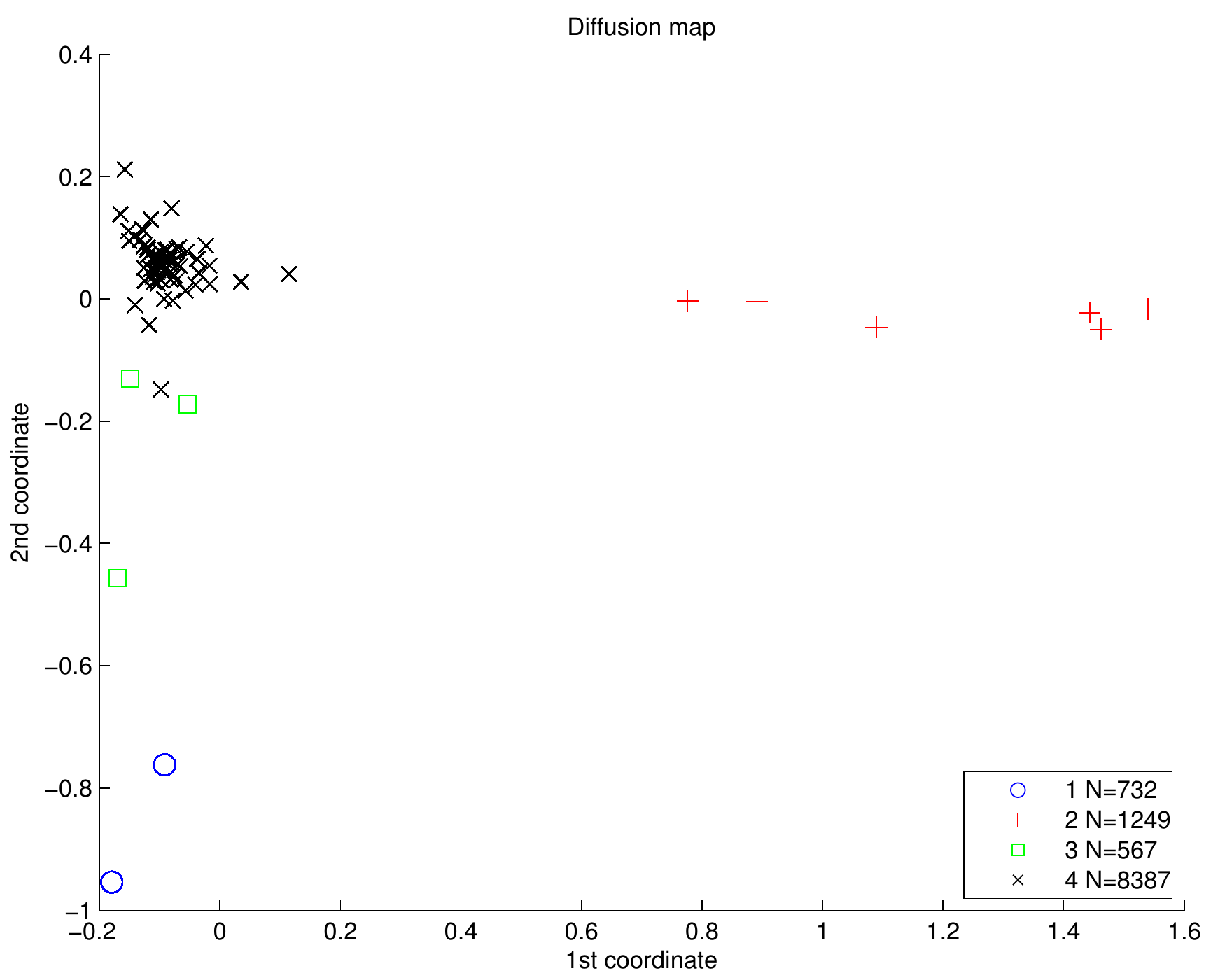}
\caption{Two-dimensional visualization of the diffusion map of the whole dataset.}
\label{fig:dm}
\end{figure}

The second dataset contains 10,935 log lines. We use the same procedure as with the smaller dataset. In this data, 414 unique bigrams are found, resulting in a matrix of size $10,935 \times 414$. For ruleset learning phase, a small number of data points are used. After dimensionality reduction, the number of clusters $k$ is determined as described in section \ref{subsec:cluster}. Figure \ref{fig:cluster_goodness_real} shows that the best number of clusters is 4. Figure \ref{fig:dm} shows all of the data points after dimensionality reduction and clustering used for unsupervised labeling step. As we can see from this visualization, cluster $c_4$ contains clearly more points than the others, which suggests that it represents the most common behavior in the data.

We randomly select training data for the rule extraction algorithm. The remaining lines are used for traffic classification testing. With 500 randomly selected training data points, the rule extraction algorithm finds a ruleset that covers the data perfectly. This might suggest overlearning. Table \ref{tab:log_500} shows these results. In order to generalize the ruleset, training with 100 randomly selected training data points was also performed. The confusion matrix in this case is shown in Table \ref{tab:log_100}. The smaller training dataset is only about 1\% of the whole data. This introduces some inaccuracies to the classification results. Because the real labeling is not known, no actual performance metrics can be calculated.

\input{log_500_train.tex}

\input{log_100_train.tex}

\subsection{Discussion of data selection}

\noindent When using real-world data for training, it must be selected carefully because it is possible to insert delusive information that will result in improper training. Therefore limits exist for automatic signature generation in this setting \citep{NEWSOME2006,CHUNG2007,VENKATARAMAN2008}. With intelligent selection of data it is possible to make the training phase more robust against attacks. One way to overcome this is to deliberately insert malicious traffic by the security experts to the training data. Alternatively, the training data can be pre-screened before training to ensure its authenticity \citep{NEWSOME2006}. 
In this case the data was as given. We had no control on the data collection and it contains real attacks.

%% file: log_500_train.tex

\begin{table}[!t]

\small
\renewcommand{\arraystretch}{1.5}
\caption{Confusion matrix for log data with 500 training samples.}
\label{tab:log_500}

\centering
\begin{tabular}{c c | c c c c }

& & \multicolumn{4}{c}{\textbf{predicted}} \\
& &  class 1 & class 2 & class 3 & class 4 \\
\hline
\multirow{4}{*}{\rotatebox{90}{\hspace{-5pt}\textbf{actual}}} 
& class 1 &         697 &          0 &          0 &          0 \\
& class 2 &           0 &       1,191 &          0 &          0 \\
& class 3 &           0 &          0 &        539 &          0 \\
& class 4 &           0 &          0 &          0 &       8,008

\end{tabular}

\end{table}

%% file: log_100_train.tex

\begin{table}[!t]

\small
\renewcommand{\arraystretch}{1.5}
\caption{Confusion matrix for log data with 100 training samples.}
\label{tab:log_100}

\centering
\begin{tabular}{c c | c c c c }

& & \multicolumn{4}{c}{\textbf{predicted}} \\
& &  class 1 & class 2 & class 3 & class 4 \\
\hline
\multirow{4}{*}{\rotatebox{90}{\hspace{-5pt}\textbf{actual}}} 
& class 1 &         725 &          0 &          0 &          0 \\
& class 2 &           0 &       1,236 &          0 &          0 \\
& class 3 &          43 &          0 &        519 &          0 \\
& class 4 &           0 &         55 &          0 &       8,257

\end{tabular}

\end{table}

%% file: conclusion.tex
\section{Conclusion}




\noindent We present an anomaly detection system based on the concept of conjunctive rules, which could be used for automated big data monitoring. In particular, we use network log data to demonstrate the feasibility of the system. To achieve this, the structure of the data is learned in an unsupervised manner using the diffusion map dimensionality reduction methodology and labeling is acquired using clustering. Finally, conjunctive rules are extracted using the learned data classification.


Performance of the presented system was tested using KDD Cup 99 dataset, and the practical feasibility was evaluated with real-world data. The results are satisfactory even with a limited amount of training data. In the KDD Cup dataset, the small random subset has a higher Matthews correlation coefficient because the small sample does not include the whole variety of attacks, while the whole 10\% dataset is more diverse. The KDD Cup dataset has its own issues, for example the high amount of attacks compared to normal traffic. This makes automatic anomaly detection very challenging. The rule learning process achieves good performance with the real-world dataset. However, because the actual attack labeling is not known, performance metrics cannot be calculated. Nonetheless, manual inspection suggests that normal and anomalous traffic are separated, and we confirm that actual intrusions are detected. 


The system is aimed to be a tool for analyzing big datasets and detecting anomalies. 
It has the following main benefits:

\begin{itemize}
\item The resulting ruleset classifies traffic very efficiently. This enables big data classification in real-time. The traffic classification phase is much faster than the preceding rule creation phase. 
\item The unsupervised learning phase enables label creation without prior information about the data.
\item The system aims to explain the classification result of a non-linear black box algorithm with rules that are easy to understand and practical for domain experts.
\item The resulting rule-based classifier is simple to use and to implement, since only specific features need to be extracted and it is easy to compare the feature vectors to the conjunctive rules. 
\end{itemize}


The main limitation of the system is the fact that training data has to represent the structure of the new incoming network traffic as accurately as possible. If the training data differs significantly from the current network data, created rules will not work in a satisfactory way. This can be prevented by using periodical training with new, more comprehensive data. Another concern is the performance of the learning phase. A bigger training set might give better results but training will take more time. However, rule creation is performed only periodically, while the fast traffic classification is done in real-time.


For future research, more testing with different kinds of data should be done in order to validate the feasibility of the rule extraction framework. So far we have used actual server HTTP log data and KDD Cup 99 dataset. The system could also be expanded to suit other cyber security and big data scenarios, not just network intrusion detection. This involves modifications to preprocessing but does not change the other components.
In addition, some parts of the algorithms can be easily parallelized. This speeds up the processing and makes the system more scalable.